\documentclass[10pt,twocolumn,letterpaper]{article}

\usepackage{iccv}
\usepackage{times}
\usepackage{graphicx}
\usepackage{amsmath}
\usepackage{amssymb}
\usepackage{color}
\usepackage{verbatim}
\usepackage{booktabs}
\usepackage{algpseudocode}
\usepackage{algorithm}
\usepackage{amscd}
\usepackage{setspace}

\newtheorem{thm}{Theorem}

\newtheorem{dfn}{Definition}
\newtheorem{prb}{Problem}

\def\A{\mathcal{A}}

\def\C{\boldsymbol{C}}
\def\D{\boldsymbol{D}}
\def\N{\mathcal{N}}
\def\X{\mathcal{X}}
\def\S{\mathcal{S}}
\def\V{\mathcal{V}}
\def\H{\mathcal{H}}
\def\I{\mathcal{I}}
\def\J{\mathcal{J}}
\def\Y{\mathcal{Y}}
\def\Z{\mathcal{Z}}
\def\0{\boldsymbol{0}}

\def\b{\boldsymbol{b}}
\def\c{\boldsymbol{c}}
\def\y{\boldsymbol{y}}
\def\x{\boldsymbol{x}}

\def\c{\boldsymbol{c}}
\def\Re{\mathbb{R}}
\def\transpose{\top}
\newcommand{\myparagraph}[1]{\smallskip\noindent\textbf{#1.}}

\newcommand{\be}{\begin{equation}}
\newcommand{\ee}{\end{equation}}
\newcommand{\bq}{\begin{eqnarray}}
\newcommand{\eq}{\end{eqnarray}}

\newcommand{\ba}{\left[ \begin{array}}
\newcommand{\ea}{\\ \end{array} \right]}

\newcommand{\ra}[1]{\renewcommand{\arraystretch}{#1}}

\graphicspath{{figures/}}

\iccvfinalcopy

\def\iccvPaperID{13} 

\begin{document}

\title{Filtrated Spectral Algebraic Subspace Clustering \thanks{This work was supported by grant NSF 1447822.}}

\author{Manolis C. Tsakiris and Ren\'e Vidal\\
Center for Imaging Science, Johns Hopkins University\\
3400 N. Charles Street, Baltimore, MD, 21218, USA\\
{\tt\small m.tsakiris,rvidal@jhu.edu}
}
\maketitle

\begin{abstract}
Algebraic Subspace Clustering (ASC) is a simple and elegant method based on polynomial fitting and differentiation for clustering noiseless data drawn from an arbitrary union of subspaces. In practice, however, ASC is limited to equi-dimensional subspaces because the estimation of the subspace dimension via algebraic methods is sensitive to noise. This paper proposes a new ASC algorithm that can handle noisy data drawn from subspaces of arbitrary dimensions. The key ideas are (1) to construct, at each point, a decreasing sequence of subspaces containing the subspace passing through that point; (2) to use the distances from any other point to each subspace in the sequence to construct a subspace clustering affinity, which is superior to alternative affinities both in theory and in practice. Experiments on the Hopkins 155 dataset demonstrate the superiority of the proposed method with respect to sparse and low rank subspace clustering methods.	
\end{abstract}

\section{Introduction}

Subspace clustering is the problem of clustering a collection of points drawn approximately from a union of linear subspaces. This is an important problem in pattern recognition with diverse applications from computer vision  \cite{Vidal:SPM11-SC} to genomics \cite{Mcwilliams14subspace}.

\myparagraph{Related Work}
Early subspace clustering methods  were based on alternating between finding the subspaces given the clustering and vice versa \cite{Bradley:JGO00,Tseng:JOTA00,Tipping-mixtures:99}, and were 
very sensitive to initialization.
The need for good initialization motivated the development of an algebraic technique called Generalized Principal Component Analysis (GPCA) \cite{Vidal:PAMI05}, which solves the problem in closed form. The key idea behind GPCA is that a union of $n$ subspaces can be represented by a collection of polynomials of degree $n$, with the property that their gradients at a data point give the normals to the subspace passing through that point. This is exploited in \cite{Vidal:CVPR04-gpca} and \cite{Huang:CVPR04-ED} for clustering a known number of subspaces. The recent Abstract Algebraic Subspace Clustering (AASC) method of \cite{Tsakiris:Asilomar14,Tsakiris:AASC15}, unifies the ideas of \cite{Huang:CVPR04-ED,Vidal:PAMI05}, into a provably correct method for
the decomposition of a union of subspaces to its constituent subspaces. However, while in theory GPCA and AASC are applicable to subspaces of any dimensions, in practice the estimation of the subspaces is sensitive to data corruptions. 

The need for methods that can handle high-dimensional data corrupted by noise and outliers motivated the quest for better subspace clustering affinities. State-of-the-art methods, such as \emph{Sparse Subspace Clustering} \cite{Elhamifar:CVPR09,Elhamifar:ICASSP10,Elhamifar:TPAMI13} and \emph{Low Rank Subspace Clustering} \cite{Liu:ICML10,Favaro:CVPR11,Vidal:PRL14,Liu:TPAMI13}, exploit the fact that a point in a union of subspaces can always be expressed as a linear combination of other points in the subspaces. Sparse and low rank representation techniques are then used to compute the coefficients, which are then used to build a subspace clustering affinity. These methods perform very well when the subspace dimensions are much smaller than the dimension of the ambient space, the subspaces are sufficiently separated and the data are well distributed inside the subspaces \cite{Elhamifar:TPAMI13,Soltanolkotabi:AS12}. However, these methods fail when the dimensions of the subspaces are large, e.g. a union of hyperplanes, which is the case where GPCA,
to be henceforth called Algebraic Subspace Clustering (ASC), performs best.
In addition, sparse methods produce low Inter-Class Connectivity, but the Intra-Class Connectivity is also low due to sparsity, leading to over-segmentation issues. Conversely, Low-Rank and $\ell_2$ methods produce high Intra-Class Connectivity (since they are less sparse) but this also leads to high Inter-Class Conectivity.
Consequently, there is a strong need for developing methods that produce high Intra-Class and low Inter-Class connectivity.

\myparagraph{Paper Contributions}
The main contribution of this paper is to propose a new subspace clustering algorithm that can handle noisy data drawn from a union of subspaces of different dimensions.
The key idea is to construct for each data point (the \emph{reference point}) a sequence of projections onto hyperplanes that contain the \emph{reference subspace} (the subspace associated to the  reference point). The norms of the projected data points are used to define their affinity with the reference point. This process leads to an affinity matrix of high intra-class and low cross-class connectivity, upon which spectral clustering is applied. We provide a theorem of correctness of the proposed algorithm in the absence of noise as well as a variation suitable for noisy data.
As a secondary contribution, we propose to replace the angle-based affinity proposed in \cite{Vidal:PAMI05} by a superior distance-based affinity. This modification is motivated by the fact that the angle-based affinity is theoretically correct only in the case of hyperplanes, and is not a good affinity for subspaces of varying dimensions. 
Our experiments demonstrate
that the proposed method outperforms other subspace clustering algorithms on the Hopkins 155 motion segmentation database as well as on synthetic experiments for arbitrary-dimensional subspaces of a low-dimensional ambient space.

\section{Algebraic Subspace Clustering: A Review} \label{section:ASC-overview}

We begin with a brief overview of the ASC theory and algorithms. We refer the reader to \cite{Vidal:CVPR03-gpca,Vidal:PAMI05,Derksen:JPAA07,Ma:SIAM08} for details.

\myparagraph{Subspace Clustering Problem} 
Let $\X = \left\{\x_1,\dots,\x_N\right\}$ be a set of points that lie in an unknown union of $n>1$ subspaces $\A=\S_1 \cup \cdots \cup \S_n$, where $\S_i$ a linear subspace of $\Re^D$ of dimension $d_i <D$. The goal of subspace clustering is to find the number of subspaces, a basis for each subspace, and cluster the data points based on their subspace membership, \ie, find the correct decomposition or \emph{clustering} of $\X$ as $\X = \X_1 \cup \cdots \cup \X_n$, where $\X_i = \X \cap \S_i$. 
To make the subspace clustering problem well-defined, we need to make certain assumptions on the geometry of both the subspaces $\S_i$ and the data $\X$. In this work we assume that the underlying union of subspaces $\A$ is \emph{transversal} \cite{Ma:SIAM08}, which in particular implies that there are no inclusions between subspaces. Moreover, we assume that $ \X_i \cap \S_{i'} = \emptyset$ for $i \neq i'$, \ie, each of the given points is associated to a unique subspace. This guarantees that the above decomposition of $\X$ is in fact a partition, and it is unique. A final assumption that we need is that the data $\X$ are rich enough and in general position (see Definition \ref{def:general-position}).


\myparagraph{Unions of Subspaces as Algebraic Varieties}
A key idea behind ASC is 
that a union of $n$ subspaces $\A=\S_1 \cup \cdots \cup \S_n$ of $\Re^{D}$ is the \emph{zero set} of a finite set of homogeneous polynomials of degree $n$ with real coefficients in $D$ indeterminates $(x_1,\dots,x_D)$. Such a set is called an \emph{algebraic variety}~\cite{AtiyahMacDonald-1994}. For example, a union of $n$ hyperplanes $\A = \H_1 \cup \cdots \cup \H_n$, where the $i$th hyperplane $\H_i = \{ \x: \b_i^\transpose\x = 0\}$ is defined by its normal vector $\b_i\in\Re^D$, is the zero set of  
\begin{equation}
\label{eq:hyperplanes}
p(x) = (\b_1^\top x) ( \b_2^\top x) \cdots (\b_n^{\transpose} x).
\end{equation}
Likewise, the union of a plane with normal $\b$ and a line with normals $\b_1,\b_2\in\Re^3$ is the zero set of the two polynomials 
$p_1(x) = (\b^{\transpose} x) (\b_1^{\transpose} x)$ and
$p_2(x) = (\b^{\transpose} x) (\b_2^{\transpose} x)$.
Observe that these \emph{vanishing polynomials} are homogeneous of degree $n$, where $n$ is the number of subspaces. Moreover, they are factorizable into linear forms, with each subspace contributing a linear form to the product. Each such linear form is in turn defined by a normal vector to the subspace.


\myparagraph{Finding Vanishing Polynomials} 
Note that the coefficients of the polynomials associated with a union of subspaces $\A$ can be obtained from sufficiently many samples $\X\subset \A$ in \emph{general position} by solving a linear system of equations. 
\begin{dfn}
\label{def:general-position}
We say that the data $\X \subset \A$ is in general position if a degree $n$ polynomial vanishes on $\X$ if and only if it vanishes on the underlying union of subspaces $\A$.
\end{dfn}
For example, if $\A$ is a union of two planes in $\Re^3$ with normals $\b_i = (b_{i1},b_{i2},b_{i3})$, $i=1,2$, then we can write $p$ as
\begin{align}
p(x) &= (b_{11} x_1 + b_{12}x_2 + b_{13} x_3) (b_{21} x_1 + b_{22}x_2 + b_{23} x_3) \nonumber\\
&= c_1 x_1^2 + c_2 x_1 x_2 + \cdots + c_6 x_3^2 = 
\c^\transpose\nu_2(x),
\end{align}
where $\c = (c_1,\dots,c_6)$ and $\nu_2(x) = (x_1^2,x_1x_2, \dots, x_3^2)$. Thus, we can find the vector of coefficients $\c$ by solving the set of linear equations $\c^\transpose \nu_2(\x_j) = 0$ for $j=1,\dots,N$. 
More generally, each polynomial of degree $n$ can be written as $p(x) = \c^\transpose \nu_n(x)$, where $\nu_n : \Re^D \to\Re^{\mathcal{M}_n(D)}$ is the Veronese embedding of degree $n$ that maps a point $\x \in \Re^D$ to all 
$\mathcal{M}_n(D):= {n+D-1 \choose n}$  distinct monomials of degree $n$ in the entries of $\x$. Consequently, a basis for the set of polynomials of degree $n$ that vanishes in $\X$ can be found by computing a basis for the right nullspace of the embedded data matrix, \ie, by solving the linear system:
\begin{equation}
\label{eq:veronese-nullspace}
\V_n(\X) \c = [\nu_n(\x_1) \,  \nu_n(\x_2) \,  \cdots \,  \nu_n(\x_N)]^\transpose\c = \0.
\end{equation}
However, the polynomials obtained by the above procedure
may not factorize into a product of linear forms because the space of  factorizable polynomials is not a linear space, e.g $(x_1+x_2)x_1 - (x_1-x_2) x_2 = x_1^2 + x_2^2$ is not factorizable. 

\myparagraph{Polynomial Differentiation Algorithm} Even though an elegant solution based on polynomial
factorization exists for the case of hyperplanes \cite{Vidal:CVPR03-gpca}, it has not been generalized
for subspaces of different dimensions.
However, an alternative solution has been achieved by observing that given \emph{any} degree $n$ vanishing polynomial $p$ on $\A$, and a point $\x$ in $\A$, the gradient of $p$ evaluated at $\x$ will be orthogonal to the subspace associated with point $\x$ (see \cite{Vidal:PAMI05} and \cite{Ma:SIAM08} for a geometric and algebraic argument respectively). Consequently, for the purpose of computing normal vectors to the subspaces, it is enough to compute general vanishing polynomials of degree $n$. The set of all such polynomials, denoted $\I_{\X,n}$, is a finite-dimensional vector space and a basis can be computed as a basis of the right nullspace of the Veronese matrix
$\V_n(\X) := [\nu_n(\x_1) \,  \nu_n(\x_2) \,  \cdots \,  \nu_n(\x_N)]^\transpose$,  where $\nu_n : \Re^D \to\Re^{\mathcal{M}_n(D)}$ is the Veronese embedding of degree $n$ that maps a point $\x \in \Re^D$ to all 
$\mathcal{M}_n(D):= {n+D-1 \choose n}$  distinct monomials of degree $n$ in the entries of $\x$.
Having a basis $p_1,\dots,p_s$ for $\I_{\X,n}$, it can be shown that the subspace associated to a point $\x \in \X$ can be identified as the orthogonal complement of the subspace spanned by the vectors $\nabla p_1|_{\x}, \nabla p_2|_{\x},\dots, \nabla p_s|_{\x}$  \cite{Vidal:PAMI05,Ma:SIAM08}. Then we can remove the points that lie in the same subspace as $\x$ and iterate the procedure with the remaining points until all subspaces have been identified. It is remarkable that this procedure is provably correct for a known number of subspaces of arbitrary dimensions.
Even though this result is general and insightful, algorithms that are directly based on it are extremely sensitive to noise. The main reason is that any procedure for estimating the dimension of the nullspace will unavoidably involve thresholding the singular values of $\V_n(\X)$, which will in turn yield very unstable estimates of the subspaces and subsequently poor clustering of the points.

 
\myparagraph{Spectral Algebraic Subspace Clustering Algorithm}
In the interest of enhancing the robustness of ASC in the presence of noise and obtaining a working algebraic algorithm, the standard practice has been to apply a variation of the polynomial differentiation algorithm based on spectral clustering. More specifically, given noisy data $\X$ lying close to a union of $n$ subspaces $\A$, one computes an approximate vanishing polynomial $p$ whose coefficients are given by the right singular vector of $V_n(\X)$ corresponding to its smallest singular value. Given $p$, one computes the gradient of $p$ at each point in $\X$ (which gives a normal vector associated with each point in $\X)$, and builds an affinity matrix between points $\x_j$ and $\x_{j'}$ as the cosine of the angle between their corresponding normal vectors, \ie,
\begin{align}
\C_{jj'} = \Big | \Big \langle \frac{\nabla p|_{\x_j}}{||\nabla p|_{\x_j}||},  \frac{\nabla p|_{\x_{j'}}}{||\nabla p|_{\x_{j'}}||} \Big \rangle \Big |. \label{e:ABA}
\end{align} 
This affinity is then used as input to any spectral clustering algorithm to obtain the clustering $\X = \cup_{i=1}^n\X_i$. We call this Spectral ASC method with \emph{angle-based affinity} as SASC-A.
To gain some intuition on $\C$, suppose $\A$ is a union of $n$ hyperplanes and that there is no noise. Then $p$ must be of the form in \eqref{eq:hyperplanes}. In that case $\C_{jj'}$ is simply the cosine of the angle between the normals to the hyperplanes that are associated with points $\x_j$ and $\x_{j'}$. If both points lie in the same hyperplane, their normals must be equal, and hence $\C_{jj'} = 1$. Otherwise, $\C_{jj'} < 1$ is the cosine of the angles between the hyperplanes. Thus, assuming that these angles are not small, and that the points are well distributed on the union of the hyperplanes, spectral clustering on the affinity matrix $\C$ will in general yield the correct clustering. 
Even though SASC-A is much more robust in the presence of noise than purely algebraic methods for the case of a union of hyperplanes, it is fundamentally limited by the fact that it applies only to unions of hyperplanes. Indeed, if the orthogonal complement of a subspace $\S$ has dimension greater than $1$, there may be points $\x, \x' $ inside $\S$ such that the angle between $\nabla p|_{\x}$ and $\nabla p|_{\x'}$ is as large as $90^\circ$. In such instances, points associated to the same subspace may be weakly connected and thus there is no guarantee for the success of spectral clustering. 

\myparagraph{Abstract Filtration Scheme} Motivated by the limitation of the polynomial differentiation algorithm to a known number of subspaces, and the association of undesired \emph{ghost-subspaces} with the recursive method of \cite{Huang:CVPR04-ED}, an alternative algebraic subspace clustering procedure based on \emph{filtrations of subspace arrangements} was proposed in \cite{Tsakiris:Asilomar14,Tsakiris:AASC15}. The procedure is abstract in the sense that it receives as input a union $\A \subset \Re^D$ of an unknown number of subspaces of arbitrary dimensions, and it decomposes it to the list of its constituent subspaces. This is done recursively by identifying a single subspace each time: $\A$ is intersected with the hyperplane $\V_1$, whose normal vector is the gradient of a vanishing polynomial at a point $\x \in \A$. Then $\V_1$ contains the subspace $\S$ associated to $\x$ and so does the new \emph{smaller} union of subspaces $\A_1=\A \cap \V_1$. Next $\A_1$ is intersected with a hyperplane $\V_2$ of $\V_1$, whose normal is the gradient of a vanishing polynomial of $\A_1$ evaluated at $\x$. As before, $\A_2 = \A_1 \cap \V_1$ contains $\S$ and the process repeats until no non-zero vanishing polynomial exists, in which case $\S$ is precisely $\V_c, \, c=D -\dim \S$. By picking a point $\x \in \A - \S$ a new subspace $\S'$ is identified and so on. This method has very strong theoretical guarantees (for noiseless data) but is fairly abstract in nature. It is the very purpose of the remaining of this paper to adapt the work of \cite{Tsakiris:Asilomar14,Tsakiris:AASC15} to a numerical algorithm and to experimentally demonstrate its merit.



\section{Filtrated Spectral ASC} \label{section:FSASC}

In this section, we propose a new subspace clustering procedure which addresses the robustness of ASC with respect to noise and unknown subspace dimensions, especially in the case of  subspaces of varying dimensions.

\subsection{A Distance-Based Affinity} \label{subsection:SASC-DBA}

Our first contribution is to replace the angle-based affinity in \eqref{e:ABA} by a distance-based-affinity and to show that the new affinity possesses superior theoretical guarantees.

Given unit norm data points $\X=\left\{\x_j\right\}_{j=1}^N$ lying close to an unknown union of $n$ subspaces, let $p$ be an approximate vanishing polynomial whose coefficients are given by the right singular vector of $V_n(\X)$ associated with its smallest singular value. We define the \emph{distance-based-affinity} as
\begin{align}
\D_{jj'} = 1 
- {1\over 2} \big |\langle \nabla p|_{\x_j},  \x_{j'} \rangle \big| 
- {1\over 2}\big |\langle \nabla p|_{\x_{j'}},  \x_{j} \rangle \big| \label{eqn:DBA}
\end{align} 
where the gradient vectors are assumed to be normalized to unit Euclidean norm. We will refer to this Spectral ASC method with the \emph{distance-based affinity} in \eqref{eqn:DBA}  SASC-D. The denomination \emph{distance-based} comes from the fact that the Euclidean distance from point $\x_{j'}$ to the hyperplane $\H^{(j)}$ defined by the unit normal vector $\nabla p|_{\x_j}$ is precisely $\big|\langle \nabla p|_{\x_j}, \x_{j'} \rangle \big|$. Moreover, $\H^{(j)}$ contains the subspace passing through $\x_j$. Thus, if $\x_j$ and $\x_{j'}$ are in the same subspace, then the distance from $\x_{j'}$ to $\H^{(j)}$ is zero and so is the distance from $\x_j$ to $\H^{(j')}$. This implies that $\D_{jj'} = 1$. Of course, it may be the case that $\D_{jj'}=1$ for points $\x_j$ and $\x_{j'}$ coming from distinct subspaces.
For instance, consider a union of two lines in $\Re^3$ and choose a plane containing one of the lines. If the plane happens to contain the two lines, then $\D_{jj'}=1$ for all pairs of points in the two lines.

%
%
\begin{thm} \label{thm:DBA}
Let $\{\x_j\}_{j=1}^N$ be points of $\Re^D$ lying in a union of $n$ subspaces $\A$. Let $p$ be a homogeneous polynomial of any degree vanishing on $\A$. Then the distance-based affinity in \eqref{eqn:DBA} is such that if points $\x_j,\x_{j'}$ lie in the same subspace, then $D_{jj'} = 1$. The converse is not true in general. 
\end{thm} 

\subsection{Filtrated ASC} \label{subsection:FiltrationScheme}

Theorem \ref{thm:DBA} shows the superiority of the distance-based affinity in \eqref{eqn:DBA} over the angle-based affinity in \eqref{e:ABA} because it ensures that points from the same subspace will be given an affinity of maximal value $1$. What still limits the theoretical guarantees of \eqref{eqn:DBA} is the fact that points from distinct subspaces may also have a maximal affinity $1$. 

In this section, we show that it is possible to further refine \eqref{eqn:DBA} by a filtration process illustrated in Figure \ref{fig:filtration}. Let $\X = \{\x_1,\dots,\x_N\}$ be a set of points in $\Re^D$ in general position in a union of $n$ transversal subspaces $\A=\cup_{i=1}^n \S_i$. Assume that each point lies in only one of the $n$ subspaces and is normalized to have unit norm. The key idea behind the filtration process is that, given an arbitrary \emph{reference point} $\x\in\X$ in one of the subspaces, say $\S$, we can identify all other data points in the same subspace as $\x$ by
1) projecting all data points in $\X$ onto $\S$ and
2) finding the points in $\X$ whose norm after projection remains equal to one.

\begin{figure}
\centering
$\minCDarrowwidth17pt
\begin{CD} 
\mathbb{R}^D @<<< \Re^{D-1} @<<< \cdots @<<< \Re^{D-c+1}@<<<\Re^{D-c} \cong \S
\\ @AAA	@AAA	 @. @ AAA @AAA \\
\X @<<< \bar{\X}_{1} @<<< \cdots @<<< \bar{\X}_{c-1} @<<<\bar{\X}_{c}
\end{CD}
$
\smallskip
\caption{Commutative diagram of the filtration associated with a reference point $\x\in\X$. The arrows denote embeddings.}
\label{fig:filtration}
\end{figure} 

The fundamental challenge, however, is that we do not know $\S$. The filtration process in Figure \ref{fig:filtration} is designed, precisely, to perform a sequence of projections, which ultimately give the projection onto $\S$ without knowing $\S$.

At step 1 of the filtration, choose a vanishing polynomial $p_0$ of $\A$ of degree $n$ from the nullspace of $\V_n(\X)$ such that $\nabla p_0 |_{\x} \neq \0$. One can show that such a $p_0$ always exists. 
Let $\b_1:= \nabla p_0 |_{\x} / ||\nabla p_0 |_{\x}||$ and let $\H_1$ be the hyperplane of $\Re^D$ defined by $\b_1$. If $| \langle \b_1, \x_j \rangle| > 0$, then by Theorem \ref{thm:DBA} we know that point $\x_j$ is not in $\S$. Consequently, we can filter the set $\X$ to obtain a subset $\X_{1} := \{ \x_j \in \X : |\langle \b_1, \x_j \rangle| = 0 \}$. Geometrically, $\X_{1}$ is precisely the subset of $\X$ that lies inside the hyperplane $\H_1$, \ie, $\X_{1} = \X \cap \H_1$. 

The key observation now is that $\X_{1}$ is a set of points of $\Re^D$ drawn from the union of subspaces $\A_1 := \A \cap \H_1  = \bigcup_{i=1}^n (\S_i \cap \H_1)$. But $\A_1$ is up to isomorphism a union of subspaces of $\Re^{D-1}$, since it is embedded in the hyperplane $\H_1$. In particular, consider the composite linear transformation $\pi_1: \Re^D \rightarrow \H_1 \xrightarrow{\sim} \Re^{D-1}$, where the first arrow is the orthogonal projection of $\Re^D$ onto $\H_1$ and the second arrow maps a basis of $\H_1$ to the standard basis of $\Re^{D-1}$. We can replace the redundant representation $\X_{1}$ by $\bar{\X}_{1} := \pi_1(\X_{1} ) \subset \Re^{D-1}$. It is important to note that the norm of every point in $\X_{1}$ remains unchanged and equal to $1$ under the transformation $\pi_1$. Note also that now $\bar{\X}_{1}$ may be actually a subset of a union of $n_1 \le n$ subspaces of $\Re^{D-1}$, as it is quite possible that all points in $\X$ lying in some subspace $\S_{i} \neq \S$ were filtered out. 

Now, since $\X$ is in general position inside $\A$, $\X_1$ will be in general position inside $\S_1$, and from this one can deduce that every vanishing polynomial of $\bar{\X}_{1}$ has gradient orthogonal to $\pi_1(\S_1)$ at $\pi(\x)$, and that  there is a vanishing polynomial $p_1$ of degree $n_1$ such that $\nabla p_1 |_{\pi_1(\x)} \neq 0$. Let $\H_2$ be a hyperplane of $\Re^{D-1}$ defined by the normal vector $\b_2 : = \nabla p_1 |_{\pi_1(\x)}/||\nabla p_1 |_{\pi_1(\x)}||$. Note that $\H_2$ contains all the points of $\bar{\X}_{1}$ that correspond to $\S$. As before, we can filter the set $\bar{\X}_{1}$ to obtain a new set 
$\X_2 = \bar{\X}_{1} \cap \H_2$. Once again, $\X_2$ lies in a union of at most $ n_1$ subspaces of $\Re^{D-1}$, which is however embedded in the hyperplane  $\H_2$, and thus we can replace $\X_2$ by its image $\bar{\X}_{2}$ under the composite linear transformation $\pi_2: \Re^{D-1} \rightarrow \H_2 \xrightarrow{\sim} \Re^{D-2}$, in which the first arrow is the orthogonal projection of $\Re^{D-1}$ onto $\H_2$, and the second arrow maps a basis of $\H_2$ to the standard basis of $\Re^{D-2}$. 
Proceeding inductively, this process will terminate precisely after $c$ steps, where $c = D - \dim(\S)$ is the codimension of $\S$. More specifically, there will be no non-zero vanishing polynomials on $\bar{\X}_c$ and $\H_{c}$ will be isomorphic to $\S$. Thus $\bar{\X}_{c}$ will consist of the images of the points of $\X \cap \S$ under the sequence of transformations $\pi_{c} \circ \pi_{c-1} \circ \cdots \circ \pi_{1}$. We note that the norm of these points remains unchanged and equal to $1$ under $\pi_{c} \circ \pi_{c-1} \circ \cdots \circ \pi_{1}$. 

Once the points of $\X$ that lie in $\S$ have been identified, we can remove them and repeat the process starting with the set $\X - \X \cap \S_1$, which lies in general position inside $\S_2\cup \cdots \cup \S_n$. This leads to Algorithm \ref{alg:FASC}, which we term \emph{Filtrated-Algebraic-Subspace-Clustering} (FASC), and is guaranteed to return the correct clustering:

\begin{algorithm} \caption{Filtrated Algebraic Subspace Clustering} \label{alg:FASC} 
\begin{algorithmic}[1]  
\Procedure{FASC}{$\mathcal{X}=\left\{\x_1,\hdots,\x_N\right\} \subset \Re^D, n$}
\State $\Y \gets \emptyset, \Z \gets \emptyset$;
\For {$i = 1 : n-1$}
\State $\X' \gets \X - \Z$; $d \gets D$; 
\State take any $\x \in \X'$, $k \gets 0$;
\While {$||\x|| = 1$} 
\State $k \gets k+1$;
\State find $p \in \N\left(\V_{\le n}(\X')\right)$ s.t. $\nabla p|_{\x} \neq 0$;
\State $ \pi_k \gets \left[\Re^d \rightarrow \langle \nabla p|_{\x} \rangle^\perp \xrightarrow{\sim} \Re^{d-1}\right]$;
\State $\X' \gets \left\{\pi_k(\y): \y \in \X', \langle \nabla p|_{\x}, \y \rangle = 0 \right\}$;
\State $\x \gets \pi_k(\x)$; $d \gets d-1$;
\EndWhile
\State $\X' \gets \left\{\x \in \X: ||\pi_{k-1} \circ \cdots \circ \pi_1(\x)||=1\right\}$;
\State $\Z \gets \Z \cup \X'$; $\Y \gets \Y \cup \left\{ (\X',d+1 \right)\}$;
\EndFor
\State $\Y \gets \Y \cup \left\{(\X - \Z,\text{rank}(\X-\Z)) \right\}$;
\State \Return $\Y$;
\EndProcedure 
\end{algorithmic} 
\end{algorithm}

\vspace{-3mm}
\begin{thm}
\label{thm:FASC} 
Let $\X = \left\{\x_j\right\}_{j=1}^N$ be points of $\Re^D$ lying in a transversal union of subspaces $\A=\cup_{i=1}^n\S_i$. Let $\X_i = \X \cap \S_i, \forall i \in [n]$. Assuming
that the points of $\X$ are in general position inside $\A$, Algorithm \ref{alg:FASC} returns
a set $\Y = \left\{(\Y_i,d_i) \right\}_{i=1}^n$ such that $\Y_i = \X_{\tau(i)}, \, 
d_i=\dim \S_{\tau(i)}, \, i = 1,\dots,n$, where $\tau$ is a permutation on $n$ symbols. 
\end{thm}

\subsection{Filtrated Spectral ASC}

Let us now consider the case where the data $\X$ are corrupted by noise. In this case, Algorithm \ref{alg:FASC} (FASC) is not applicable because the noisy embedded data matrix $V_n(\X)$ is in general full rank. Nonetheless, we will show next that we can still exploit the insights revealed by the theoretical guarantees of FASC to construct a Robust-ASC algorithm.

To begin with, note that Algorithm \ref{alg:FASC} requires a single vanishing polynomial at each step of each filtration. We can use any approximate vanishing polynomial at step $k$. For example, letting $\bar \X_{k-1}$ be the points that have passed through the filtration at step $k-1$, we can let $p_{k-1}$ be the polynomial whose coefficients are given by the right singular vector of $V_n(\bar \X_{k-1})$ corresponding to its smallest singular value. Notice that no thresholding is required to choose such a $p_{k-1}$. This is in sharp contrast to the polynomial differentiation algorithm described in Section \ref{section:ASC-overview}, which requires a thresholding on the singular values of $V_n(\X)$ in order to estimate a basis of $\I_{\X,n}$. Now, for any point $\x \in \X$, $p_{k-1}$ gives a hyperplane $\langle \nabla p_{k-1} |_{\x} \rangle^\perp$ that approximately contains the subspace associated to point $\x$. However, we cannot go to the next step due to the following problems.

\begin{prb} \label{prb:WhichPoint}
In general, two points lying approximately in the same subspace $\S$ will produce different hyperplanes that approximately contain $\S$ with different levels of accuracy. In the noiseless case any point would be equally good. In the presence of noise though, the choice of the reference point $\x$ becomes significant. 
How should $\x$ be chosen?
\end{prb}

\begin{prb}\label{prb:threshold}
Given a hyperplane produced by a point $\x$, we need to determine which other points in $\X$ lie approximately in the hyperplane and filter out the remaining points. A simple approach is to filter out a point if its distance to the hyperplane is above a threshold $\delta$, or 
if the relative change in its norm is more than $\delta$. Clearly the choice of $\delta$ will affect the performance of the algorithm. How should $\delta$ be chosen?
\end{prb}

\begin{prb} \label{prb:codimension}
Finally, we also need to determine the number of steps needed to stop the filtration. This is equivalent to determining the codimension of the subspace associated to the reference point of that filtration. In the noiseless case, one stops when the norm of the reference point becomes less than $1$. In the noisy case, because the hyperplanes used to construct the filtration are only approximate, the norm of the reference point could drop at every step of the filtration. 
Hence a suitable stopping criterion needs to be devised.
\end{prb}

Inspired be the SASC-D algorithm, which handles noise by computing a normal vector for each data point and uses the normal vectors to define a distance-based affinity, we propose to address Problem \ref{prb:WhichPoint} by constructing a filtration for each data point $\x_j\in\X$ with reference point $\x_j$ and using the norms of the data points to construct the affinity. 

Let $\pi_k^{(j)}$ be the projection at step $k$ of the filtration for point $\x_j$. 
Recall that at step $k$, only a subset of the original points will remain, while others will be filtered out. We can define an affinity matrix as
\begin{align}
\C^k_{jj'} = \begin{cases}
\| \pi_k^{(j)} \circ \cdots \circ \pi_1^{(j)} (\x_{j'}) \|  & \text{if $\x_{j'}$ remains}\\
0 &  \text{otherwise}.\!
\end{cases}
\end{align} 
This affinity captures the fact that if points $\x_j$ and $\x_{j'}$ are in the same subspace, then the norm of $\x_{j'}$ should not change from step $0$ to step $k$ of the filtration computed with reference point $\x_j$. Otherwise, if $\x_j$ and $\x_{j'}$ are in different subspaces, the norm of $\x_{j'}$ is expected to be reduced by the time the filtration reaches step $c=D - \dim (\S)$, where $\S$ is the reference subspace associated to $\x_j$. In the case of noiseless data, only the points in the correct subspace survive step $c$ and their norms are precisely equal to one. Therefore, $\C^{(c)}_{jj'} =1$ if points $\x_j$ and $\x_{j'}$ are in the same subspace and $\C^{(c)}_{jj'} = 0$ otherwise. In the case of noisy data, the above affinity will not be perfect due to Problems \ref{prb:threshold} and \ref{prb:codimension}, which we address next.

To address Problem \ref{prb:threshold}, let $p$ be the approximate vanishing polynomial whose coefficients are the right singular vector of $V_n(\X)$ corresponding to the smallest singular value. Let
\begin{align}
\beta(\X)  = \frac{1}{N} \sum_{j=1}^N \Big | \big \langle \x_j, \frac{\nabla p|_{\x_j}}{||\nabla p|_{\x_j}||} \big \rangle  \Big|.
\end{align} 
Notice that $\beta=0$ in the noiseless case. In the presence of noise, $\beta(\X)$ is the average over all points of the distance of a point from the hyperplane that it produces. Evidently, small levels of noise will correspond to small values of $\beta(\X)$. Thus, we propose to define $\delta = \gamma \cdot \beta(\X)$, where $\gamma >0$ is a user defined parameter. To determine $\gamma$, we propose to construct multiple filtrations for different values $\gamma_1, \dots, \gamma_M$ of $\gamma$. Each filtration will result in a different affinity matrix. Suppose  we have defined a stopping criterion to terminate each filtration so that we can use the affinity matrix at the last step of the filtration (see below for stopping criteria). Given these affinity matrices, we choose the one whose normalized Laplacian has the largest eigengap $\lambda_{n+1} - \lambda_n$, where the eigenvalues are ordered increasingly. 

To address Problem \ref{prb:codimension}, we stop the filtration at step $k$ if 
1) the number of points is less than the ambient dimension of the Veronese-embedded points;
2) the reference point $\x_j$ is filtered out at the $(k+1)$th step; or 
3) the number of points that passed through the filtration at step $k+1$ is less than some integer $\mu$. 
This integer is the smallest number of points that our algorithm is allowed to consider as a cluster. 

Finally, the resulting affinity is symmetrized, and used for spectral clustering, as described in Algorithm \ref{alg:FSASC}.\footnote{$\textsc{Spectrum}\big(NL(\C+\C^{\transpose})\big)$ denotes the spectrum of the normalized Laplacian matrix of $\C + \C^{\transpose}$, $\textsc{SpecClust}\big(\C^*+(\C^*)^\transpose,n\big)$ denotes spectral clustering being applied to $\C^*+\C^{*\transpose}$ to obtain $n$ clusters, and $\textsc{Vanishing}\big(V_n(\X)\big)$ is the polynomial whose coefficients are the right singular vector of $V_n(\X)$ corresponding to the smallest singular value.} 


\begin{algorithm} \caption{Filtrated Spectral ASC} \label{alg:FSASC} 
\begin{spacing}{0.8}
\begin{algorithmic} [1]
\Procedure{FSASC}{$\mathcal{X}, D, n,\mu,\{\gamma_m\}_{m=1}^M$}
\If {$N < \mathcal{M}_n(D)$}
\State \Return('Not enough points');
\Else
\State eigengap $\gets 0$; $\C^* \gets 0_{N \times N}$;
\State $\x_j \gets \x_j / ||\x_j||, \, \forall j \in [N]$;
\State $p \gets \Call{Vanishing}{\V_n(\X)}$;
\State $\beta \gets  \frac{1}{N} \sum_{j=1}^N \big|\langle \x_j, \frac{\nabla p|_{\x_j}}{||\nabla p|_{\x_j}||}\rangle\big|$;
\For {k = 1 : M}
	\State $\delta \gets  \beta \cdot \gamma_k, \, \C \gets 0_{N \times N}$;
	\For {j = 1 : N}
		\State $C_{j,:} \gets \Call{Filtration}{\X,\x_j,p,\mu,\delta,n}$;
	\EndFor
	\State $\{\lambda_s\}_{s=1}^N \gets \Call{Spectrum}{NL(\C+\C^\transpose)}$ ;
    \If {(eigengap $< \lambda_{n+1} - \lambda_n$)}
		\State eigengap $\gets \lambda_{n+1} - \lambda_n$; $\C^* \gets \C$;   
	\EndIf     
\EndFor
\State $\left\{\Y_i\right\}_{i=1}^n \gets \Call{SpecClust}{\C^*+\C^{*\transpose},n}$; 
\State \Return $\left\{\Y_i\right\}_{i=1}^n$;
\EndIf
\EndProcedure 
\Statex
\Function{Filtration}{$\X,\x,p,\mu,\delta,n$}

\State $d \gets D, \, \J \gets [N],  q \gets p, \boldsymbol{c} \gets 0_{1 \times N}$;
        \State  flag $\gets 1$;
			\While {($d >1$) and ($\text{flag}=1$)}						
				\State $\H \gets \langle \nabla q|_{\x} \rangle^\perp, \, \pi \gets \left[ \Re^d \rightarrow  \H \xrightarrow{\sim} \Re^{d-1} \right]$;				
				\If {$(||\x|| - || \pi(\x)||)/||\x|| > \delta$}
					\If {$d = D$}
						\State $\boldsymbol{c}(j') \gets || \pi(\x_j')||, \, \forall j' \in [N]$;						
					\EndIf			
					\State flag $\gets 0$;		
				\Else
					\State $\J \gets \left\{j' \in [N] : \frac{||\x_{j'}|| - || \pi(\x_{j'})||}{||\x_{j'}||} \le \delta \right\}$
					\If {$\left| \J \right| < \mu$}
						\State flag $\gets 0$;
					\Else
						\State $\boldsymbol{c}(j') \gets || \pi(\x_j')||, \, \forall j' \in \J$;
						\State $\boldsymbol{c}(j') \gets 0, \, \forall j' \in [N]- \J$;
						\If {$|\J| < \mathcal{M}_n(d)$}
							\State flag $\gets 0$;
						\Else
							\State $d \gets d -1, \x \gets \pi(\x)$;
                            \State $\x_{j'} \gets \pi(\x_{j'}) \, \forall j' \in \J$;
						    \State $ \X \gets \left\{\x_{j'}: j' \in \J \right\}$;											\State $q \gets \Call{Vanishing}{\V_n(\X)}$ ;
						\EndIf
					\EndIf
				\EndIf
			\EndWhile

\State \Return($\boldsymbol{c}$);
\EndFunction

\end{algorithmic} 
\end{spacing}
\end{algorithm} 

\vspace{-0.5mm}
\section{Experiments}
\vspace{-1mm}
\myparagraph{Synthetic Data} 
We randomly generate $n=3$ subspaces of dimensions $d_i \in \{1,2,3,4\}$  in $\Re^5$. For each choice of $\{d_i\}$, we randomly generate $N_i = 100$ unit norm points per subspace and add zero-mean Gaussian noise with standard deviation $\sigma\in\{0,0.01, 0.03, 0.05\}$ in the direction orthogonal to the subspace. For each choice of $\{d_i\}$ and $\sigma$, we perform $500$ independent subspace clustering experiments using the algebraic methods FSASC, SASC-D and SASC-A, and compare to state of the art methods such as SSC \cite{Elhamifar:TPAMI13}, LRR \cite{Liu:TPAMI13,Liu:ICML10}, LRSC 
\cite{Vidal:PRL14} and LSR using equation (16) in \cite{Lu:ECCV12}. We also use the heuristic post processing of the affinity for LRR (LRR-H) and LSR (LSR-H). For FSASC we use $\mu=10$ and $\gamma\in\{0.001,0.005,0.01,0.05,0.1,0.5,1,5,10\}$, for SSC $\alpha=20$ and $\rho=0.7$, for LRR $\lambda=4$, for LRSC $\tau=420, \, \alpha=4000$ and for LSR $\lambda=0.0048$. We report average clustering errors, intra-cluster connectivities of the affinity matrices $\C$ produced by the methods (defined to be the minimum \emph{algebraic connectivity}\footnote{The algebraic connectivity of a graph is the second smallest eigenvalue of the Laplacian of the graph. Here we use the normalized Laplacian.} among the subgraphs corresponding to each of the three subspaces) and inter-cluster connectivities ($\sum_{\x_j \in \S_i, \x_{j'} \in \S_{i'}, i \neq i'} |\C_{j,j'}|/||\C||_1$). Due to lack of space, we report errors on all methods only for $0 \%$ and $5 \%$ noise. 

Table \ref{table:synthetic-error} reports the mean clustering errors. Observe that FSASC is the only method that gives $0$ error for noiseless data for all dimension configurations, thus verifying experimentally its strong theoretical guarantees: no restrictions on the dimensions of subspaces are required for correctness. As expected, SSC, LRR, LRSC and LSR yield perfect clustering when $d/D$ is small, but their performance degrades significantly for large $d/D$. Observe also that, although SASC-D is much simpler than FSASC and has similar complexity to SASC-A, its performance is very close to that of FSASC, and much better than SASC-A. We attribute this phenomenon to the correctness Theorem \ref{thm:DBA} of SASC-D.
As the noise level increases, FSASC remains stable across all dimension configurations with superior behavior among all compared methods.  SASC-D is less robust in the presence of noise, except for the case of hyperplanes, in which it is the best  method. This phenomenon is expected, since SASC-D is essentially equivalent to FSASC if the latter is configured to take only one step in each filtration in Figure \ref{fig:filtration}, and this is precisely the optimal stopping point in every filtration when the subspaces are hyperplanes. In this case, if data are noisy, the criterion for stopping FSASC filtrations is determined by the parameter $\gamma$ and by the level of noise via the quantity $\beta$, leading to suboptimal values (\ie, more than one step may be taken in the filtration).

Tables  \ref{table:synthetic-intra} and \ref{table:synthetic-inter} indicate that FSASC yields higher quality affinity graphs for the purpose of clustering. To see why this is the case, observe that except for FSASC, we can distinguish two kinds of behavior in the remaining methods: the first kind gives high intra-cluster connectivity at the cost of high inter-cluster connectivity. Such methods are SASC-D, SASC-A, LRR, LRSC and LSR. The second kind gives low inter-cluster connectivity at the expense of low intra-cluster connectivity leading to unstable clustering results by the spectral clustering method. Such methods are SSC, LRR-H and LSR-H. This is expected because these methods use sparse affinities. On the other hand, FSASC circumvents this trade-off by giving high intra-cluster connectivity and low inter-cluster connectivity, thus enhancing the success of the spectral clustering step. 

\begin{table}
\centering
\small
\caption{Mean clustering error in $\%$ over $500$ independent experiments on synthetic data for $3$ subspaces of $\Re^5$ of varying dimensions $(d_1,d_2,d_3)$ and varying levels of noise $\sigma \in \{0,1,3,5\}\%$.}
 \label{table:synthetic-error}
\ra{0.7}
\begin{tabular}{@{}l@{\,}c@{\,}c@{\,}c@{\,}c@{\,}c@{\,}c@{\,}c@{\,}c@{}}\toprule[1pt]{}
method &  $(1,1,1)$ & $(2,2,2)$ & $(3,3,3)$ &  $(4,4,4)$   & $(1,2,3)$ &  $(2,3,4)$\\ \midrule[0.5pt]
$\sigma=0\%$ \\
FSASC      &     $\boldsymbol{0.00}$ & $\boldsymbol{0.00}$ & $\boldsymbol{0.00}$ & $\boldsymbol{0.00}$ & $\boldsymbol{0.00}$ & $\boldsymbol{0.00}$          \\
SASC-D        &     $0.76$ & $0.96$ & $0.03$ & $\boldsymbol{0.00}$ & $0.42$ & $0.13$        \\
SASC-A        &     $\boldsymbol{0.00}$ & $34.2$ & $23.3$ & $\boldsymbol{0.00}$ & $20.8$ & $10.7$           \\
SSC              &      $\boldsymbol{0.00}$ & $0.16$ & $6.18$ & $43.2$ & $1.38$ & $47.1$       \\              
LRR        &       $\boldsymbol{0.00}$ & $1.37$ & $25.5$ & $48.7$ & $1.84$ & $25.9$        \\
LRR-H         &       $\boldsymbol{0.00}$ & $1.15$ & $18.1$ & $48.4$ & $1.53$ & $20.8$        \\
LRSC        &       $\boldsymbol{0.00}$ & $1.32$ & $25.6$ & $48.7$ & $2.27$ & $26.5$        \\
LSR              &      $\boldsymbol{0.00}$ & $7.36$ & $34.9$ & $48.8$ & $5.42$ & $28.8$        \\
LSR-H              &      $\boldsymbol{0.00}$ & $1.31$ & $28.8$ & $29.3$ & $1.46$ & $32.7$        \\
\midrule[0.5pt]
$\sigma=1\%$\\
FSASC      &     $1.70$ & $\boldsymbol{0.20}$ & $\boldsymbol{0.22}$ & $3.17$ & $\boldsymbol{0.94}$ & $\boldsymbol{0.81}$          \\
SASC-D        &     $2.91$ & $0.58$ & $0.96$ & $\boldsymbol{2.39}$ & $1.78$ & $1.89$        \\
SSC              &      $\boldsymbol{1.51}$ & $1.27$ & $6.24$ & $42.5$ & $8.71$ & $44.2$       \\              
\midrule[0.5pt]
$\sigma=3\%$\\
FSASC      &     $4.39$ & $\boldsymbol{1.16}$ & $\boldsymbol{1.40}$ & $7.67$ & $\boldsymbol{2.82}$ & $\boldsymbol{2.88}$          \\
SASC-D        &     $8.41$ & $2.68$ & $4.05$ & $\boldsymbol{6.15}$ & $5.40$ & $6.46$        \\
SSC              &      $\boldsymbol{3.90}$ & $5.17$ & $7.35$ & $44.7$ & $9.50$ & $33.2$       \\              
\midrule[0.5pt]
$\sigma=5\%$\\
FSASC      &     $7.02$ & $\boldsymbol{2.69}$ & $\boldsymbol{3.42}$ & $11.34$ & $\boldsymbol{5.13}$ & $\boldsymbol{5.49}$          \\
SASC-D        &     $13.9$ & $6.81$ & $8.32$ & $\boldsymbol{9.98}$ & $9.98$ & $10.9$        \\
SASC-A      &     $43.6$ & $47.0$ & $32.7$ & $12.2$ & $41.9$ & $27.9$           \\
SSC            &     $\boldsymbol{6.23}$ & $10.2$ & $13.2$ & $45.9$ & $12.1$ & $35.4$       \\       
LRR         	&      $6.74$ & $4.88$ & $26.2$ & $48.8$ & $7.70$ & $25.5$        \\       
LRR-H        &      $6.37$ & $3.90$ & $20.9$ & $48.3$ & $6.73$ & $23.6$        \\
LRSC         	&      $6.32$ & $5.24$ & $26.5$ & $48.8$ & $7.93$ & $27.3$        \\
LSR            &      $9.38$ & $16.0$ & $34.3$ & $49.1$ & $17.4$ & $30.5$        \\
LSR-H        &      $10.1$ & $13.5$ & $28.8$ & $28.9$ & $10.3$ & $32.5$        \\
\bottomrule[1pt]
\end{tabular}
\end{table}

\begin{table}
\centering
\small
\caption{Mean intra-cluster connectivity in $\%$ for $3$ synthetic subspaces of $\Re^5$ of dimensions $(d_1,d_2,d_3)$ and noise $\sigma$.}
\ra{0.7}
\begin{tabular}{@{}l@{\,}c@{\,}c@{\,}c@{\,}c@{\,}c@{\,}c@{\,}c@{\,}c@{}}\toprule[1pt]{}
 method &  $(1,1,1)$ & $(2,2,2)$ & $(3,3,3)$ &  $(4,4,4)$   & $(1,2,3)$ &  $(2,3,4)$\\ \midrule[0.5pt]
$\sigma=0\%$ \\
FSASC      &     $100$ & $100$ & $100$ & $100$ & $100$ & $100$          \\
SASC-D        &     $100$ & $100$ & $100$ & $100$ & $100$ & $100$        \\
SASC-A        &     $100$ & $16.6$ & $17.7$ & $100$ & $18.2$ & $16.1$   \\
SSC              &      $40.2$ & $0.18$ & $0.40$ & $0.00$ & $0.24$ & $0.15$       \\              
LRR         &       $100$ & $42.0$ & $38.1$ & $41.6$ & $38.4$ & $31.6$        \\
LRR-H         &       $100$ & $22.2$ & $21.5$ & $21.5$ & $22.9$ & $19.8$        \\
LRSC         &       $100$ & $42.2$ & $38.0$ & $41.0$ & $38.4$ & $31.8$        \\
LSR              &      $100$ & $42.3$ & $37.9$ & $40.9$ & $37.9$ & $32.0$        \\
LSR-H              &      $100$ & $3.36$ & $0.92$ & $1.41$ & $2.17$ & $0.39$        \\
\midrule[0.5pt]
$\sigma=1\%$ \\
FSASC      &     $2.80$ & $21.5$ & $40.8$ & $46.3$ & $73.8$ & $32.8$          \\
SASC-D        &     $63.3$ & $81.8$ & $82.0$ & $77.9$ & $71.2$ & $77.2$        \\
SSC              &      $0.22$ & $0.25$ & $0.42$ & $0.00$ & $0.19$ & $0.18$       \\              
\bottomrule[1pt]
\end{tabular}
 \label{table:synthetic-intra}
\bigskip
\centering
\small
\caption{Mean inter-cluster connectivity (\%) for $3$ synthetic subspaces of $\Re^5$ of dimensions $(d_1,d_2,d_3)$, and noise $\sigma$.}
\ra{0.7}
\begin{tabular}{@{}l@{\,}c@{\,}c@{\,}c@{\,}c@{\,}c@{\,}c@{\,}c@{\,}c@{}}\toprule[1pt]{}
 method &  $(1,1,1)$ & $(2,2,2)$ & $(3,3,3)$ &  $(4,4,4)$   & $(1,2,3)$ &  $(2,3,4)$\\ \midrule[0.5pt]
$\sigma=0\%$ \\
FSASC      &     $0.0$ & $0.0$ & $0.0$ & $1.3$ & $0.0$ & $0.0$          \\
SASC-D        &     $57$ & $56$ & $55$ & $55$ & $56$ & $55$        \\
SASC-A       &  $39$ & $46$ & $39$ & $27$ & $40$ & $31$   \\
SSC              &      $0.0$ & $0.0$ & $2.2$ & $27$ & $0.0$ & $2.2$       \\         
LRR         &       $0.0$ & $30$ & $52$ & $61$ & $28$ & $50$        \\     
LRR-H         &       $0.0$ & $6.1$ & $30$ & $50$ & $3.7$ & $26$        \\
LRSC         &       $0.0$ & $31$ & $52$ & $61$ & $28$ & $51$        \\     
LSR             &      $0.0$ & $21$ & $38$ & $45$ & $19$ & $37$        \\
LSR-H              &      $0.0$ & $0.0$ & $0.0$ & $0.0$ & $0.0$ & $0.0$        \\
\midrule[0.5pt]
$\sigma=1\%$ \\
FSASC      &     $8.6$ & $4.7$ & $11$ & $28$ & $9.1$ & $20$          \\
SASC-D        &     $60$ & $58$ & $58$ & $57$ & $59$ & $58$        \\
SSC              &      $0.5$ & $0.2$ & $2.5$ & $26$ & $0.5$ & $2.9$       \\              
\bottomrule[1pt]
\end{tabular}
\label{table:synthetic-inter}
\end{table}

\myparagraph{Motion Segmentation} 
We evaluate different methods on the Hopkins155 motion segmentation data set \cite{Tron:CVPR07}, which contains 155 videos of $n=2$,$3$ moving objects, each one with $N=100$-$500$ feature point trajectories of dimension $D=56$-$80$. While SSC, LRR, LRSC and LSR can operate directly on the raw data, algebraic methods require $\mathcal{M}_n(D) \leq N$. Hence, for algebraic methods, we project the raw data onto the subspace spanned by their $D$ principal components, where $D$ is the largest integer $\le 8$ such that  $\mathcal{M}_n(D) \leq N$, and then normalize each point to have unit norm. We apply SSC to i) the raw data (SSC-raw) and ii) the raw points projected onto their first $8$ principal components and normalized to unit norm (SSC-proj). For FSASC, LRR, LRSC and LSR we use the same parameters as before, while for SSC the parameters are $\alpha = 800$ and $\rho = 0.7$. 

\begin{figure} 
\centering
\vspace*{-0.8cm}
\includegraphics[width=1\linewidth]{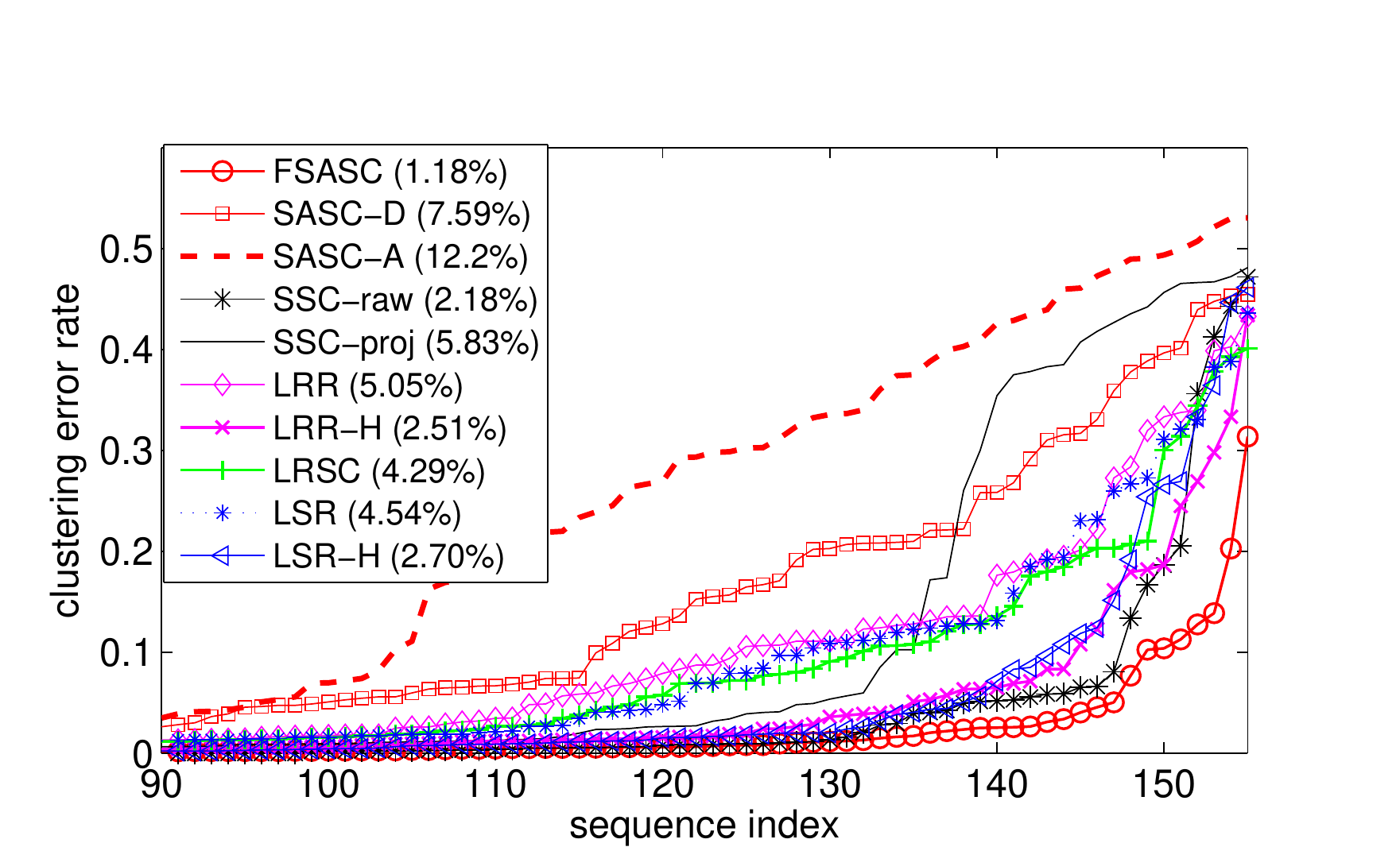}
   \caption{Clustering error ratios for both $2$ and $3$ motions in Hopkins155, ordered increasingly for each method. Errors start from the $90$-th smallest error of each method.}\label{figure:Hopkins155-ordered-error}
\end{figure}

The clustering errors and the intra/inter-cluster connectivities are reported in Table \ref{table:Hopkins155} and Fig. \ref{figure:Hopkins155-ordered-error}. Notice the clustering errors of about 5\% and 37\% for SASC-A, which is the classical GPCA algorithm. Notice how changing the angle-based by the distance-based affinity (SASC-D) already gives errors of around 5.5\% and 14\%. But most dramatically, notice how FSASC further reduces those errors to 0.8\% and 2.48\%. This clearly demonstrates the advantage of FSASC over classical ASC. Moreover, even though the dimensions of the subspaces ($d_i \in \{1,2,3,4\}$ for motion segmentation) are low relative to the ambient space dimension ($D=56$-$80$) - a case that is specifically suited for SSC, LRR, LRSC, LSR - projecting the data to $D\leq 8$, which makes the subspace dimensions comparable to the ambient dimension, is sufficient for FSASC to get superior performance relative to the best performing algorithms on Hopkins 155. We believe that this is because, overall, FSASC produces a much higher inter-cluster connectivity, without increasing the intra-cluster connectivity too much.

\setlength{\tabcolsep}{0.2em}
\begin{table}
\centering
\small
\ra{0.7}
\caption{Mean clustering error ($E$), intra-cluster connectivity ($C_1$) and inter-cluster connectivity ($C_2$) in $\%$ on the Hopkins155 data.} \label{table:Hopkins155}
\begin{tabular}
{lccrrccrrrcrrr}\toprule[1pt]
\phantom{abc}&& \multicolumn{3}{c}{$2$ motions} & \phantom{ab}& \multicolumn{3}{c}{$3$ motions} & 
\phantom{ab} & \multicolumn{3}{c}{all motions} & \\
\cmidrule{3-5} \cmidrule{7-9} \cmidrule{11-13}
method &&  $E$ &  $C_1$ & $C_2$ &&  $E$ &  $C_1$ & $C_2$ && $E$ &  $C_1$ & $C_2$  \\ \midrule
FSASC && $\boldsymbol{0.80}$ &    $18$ & $4$    && $\boldsymbol{2.48}$ &  $10$ &    $10$     
 &&  $\boldsymbol{1.18}$ &  $16$ &   $5$         \\
SASC-D   && $5.65$ &      $82$ & $26$ && $14.0$ &   $80$ & $46$  && $7.59$  &   $81$ &  $31$      \\
SASC-A    && $4.99$ &       $35$ & $5$ && $36.8$ &  $9$ & $35$ && $12.2$ &   $29$ &  $12$      \\
SSC-raw    && $1.53$ &     $5$ &  $2$   && $4.40$ &  $4$ &  $3$  &&  $2.18$ &   $5$ &  $2$       \\
SSC-proj   &&  $5.87$ & $4$ & $3$ && $5.70$ &  $3$ & 3 &&   $5.83$ &   $3$ & $3$     \\
LRR         && $4.26$ &       $25$ & $19$   && $7.78$ &  $25$ & $28$   &&   $5.05$ &   $25$ &   $21$      \\
LRR-H  && $2.25$ &  $5$ & $2$ && $3.40$ &  $4$ & $3$    &&   $2.51$ &   $5$ & $2$ \\
LRSC  && $3.38$ &  $25$ & $19$ && $7.42$ &  $24$ & $28$    &&   $4.29$ &   $25$ & $21$ \\
LSR         && $3.60$ & $24$ &  $18$  && $7.77$ &  $23$ &  $28$   &&  $ 4.54$ &    $23$ &   $21$     \\
LSR-H && $2.73$ &   $4$ &  $1$   && $2.60$ &  $3$ &  $2$   &&  $2.70$ &   $4$ &   $1$     \\
\bottomrule[1pt]
\end{tabular}
\end{table}

\myparagraph{Handwritten Digit Clustering}
In this section we consider the problem of clustering two digits, one of which is the digit $1$ (see Benford's law, e.g. \cite{Kossovsky14}).  For each pair $(1,i), \, i = 0,2,...,9,$ we randomly select $200$ images from the MNIST database  \cite{LeCun:1998} corresponding to each digit and compute the clustering errors averaged over $100$ independent experiments. SSC, LRR and LSR operate on raw data. LRR and LSR parameters are the same as before. For FSASC we set $\mu=10$ and $\gamma=1$ and for SSC we set $\alpha=10$ and $\rho=0.7$, as before. For the three algebraic methods we first project the raw data onto their first $13$ principal components and then normalize each point to have unit norm. For comparison, we also run SSC on the projected data. 
Mean errors are reported in Table \ref{table:MNIST} with SASC-A, LRR, LRSC omitted since they perform poorly (with LRR performing worse with the post-processing). We also do not show the numbers for SSC-proj since they are very close to those of SSC-raw. As in the case of motion segmentation, we observe that FSASC outperforms SASC-D (this time by a large margin), which in turn significantly outperforms SASC-A. This confirms the superiority of FSASC over previous algebraic methods. As before, FSASC is also superior to SSC. The only method that performs better is LSR-H.
We note that projecting the $784$-dimensional data onto dimension $13$, reduces the angles between the subspaces, thus
making the clustering problem harder. As a result, for more than $2$ digits the performance 
of FSASC degrades singinficantly,
even for projection dimensions up to $D=17$,
since it becomes harder for the method to distinguish the subspaces. To circumvent this issue, a higher projection dimension would be required, which currently can not be handled by FSASC,
due to the high complexity.

\begin{table}
\small
\centering
\ra{0.7}
\caption{Clustering error ($\%$) for two digits $(1,i),\, i=0,2,...,9$ in the MNIST dataset.}\label{table:MNIST}
\begin{tabular}{@{}l@{\,}c@{\,}c@{\,}c@{\,}c@{\,}c@{\,}c@{\,}c@{\,}c@{\,}c@{}}\toprule[1pt]{}
& \multicolumn{8}{c}{Digits Pair} & \phantom{ab}\\
\cmidrule{2-10}
method &  $(1,0)$ & $(1,2)$ & $(1,3)$ & $(1,4)$ & $(1,5)$   & $(1,6)$ &  $(1,7)$ & $(1,8)$ & $(1,9)$ \\ \midrule
FSASC & $\boldsymbol{0.50}$ & $4.67$ &    $1.55$     & $3.31$ &  $1.11$        
 &  $1.62$ &  $2.27$   &  $4.88$  &  $1.81$       \\
SASC-D & $4.91$ & $14.2$ & $10.3$ & $23.9$ & $8.55$ & $13.1$ & $10.2$ & $21.5$ & $17.5$       \\
SSC-raw & $1.12$ & $9.15$ & $2.66$ & $5.77$ & $2.78$ & $1.87$ & $2.90$ & $13.3$ & $2.00$       \\
LSR & $1.03$ & $5.26$ & $2.13$ & $19.1$ & $1.29$ & $1.50$ & $5.40$ & $15.3$ & $5.90$       \\
LSR-H & $0.74$ & $\boldsymbol{1.35}$ & $\boldsymbol{1.12}$ & $\boldsymbol{3.15}$ & $\boldsymbol{0.88}$ & $\boldsymbol{0.90}$ & $\boldsymbol{1.33}$ & $\boldsymbol{4.38}$ & $\boldsymbol{1.11}$       \\
\bottomrule[1pt]
\end{tabular}
\end{table}

\section{Conclusions}
\vspace{-1mm}
We presented a novel algebraic subspace clustering method based on the geometric idea of filtrations and we experimentally demonstrated its robustness to noise using synthetic and real data and its superiority to the state-of-the-art algorithms on several occasions. Overall, the method works very well for subspaces of arbitrary dimensions in a low-dimensional ambient space, and it can handle higher dimensions via a projection. The main weakness of the method is its high computational complexity, which comes from the large number of filtrations required, as well as the exponential cost of fitting polynomials to $n$ subspaces. Future research will be concerned with reducing the complexity, as well as dealing with outliers and missing entries.

\clearpage
{\small
\bibliographystyle{ieee}
\bibliography{biblio/vidal,biblio/vision,biblio/math,biblio/learning,biblio/sparse,biblio/geometry,biblio/dti,biblio/recognition,biblio/surgery,biblio/coding,biblio/matrixcompletion,biblio/segmentation}}

\begin{thebibliography}{10}\itemsep=-1pt

\bibitem{AtiyahMacDonald-1994}
M.~Atiyah and I.~MacDonald.
\newblock {\em Introduction to Commutative Algebra}.
\newblock Westview Press, 1994.

\bibitem{Bradley:JGO00}
P.~S. Bradley and O.~L. Mangasarian.
\newblock k-plane clustering.
\newblock {\em Journal of Global Optimization}, 16(1):23--32, 2000.

\bibitem{Derksen:JPAA07}
H.~Derksen.
\newblock Hilbert series of subspace arrangements.
\newblock {\em Journal of Pure and Applied Algebra}, 209(1):91--98, 2007.

\bibitem{Elhamifar:CVPR09}
E.~Elhamifar and R.~Vidal.
\newblock Sparse subspace clustering.
\newblock In {\em {IEEE} Conference on Computer Vision and Pattern
  Recognition}, 2009.

\bibitem{Elhamifar:ICASSP10}
E.~Elhamifar and R.~Vidal.
\newblock Clustering disjoint subspaces via sparse representation.
\newblock In {\em {IEEE} International Conference on Acoustics, Speech, and
  Signal Processing}, 2010.

\bibitem{Elhamifar:TPAMI13}
E.~Elhamifar and R.~Vidal.
\newblock Sparse subspace clustering: Algorithm, theory, and applications.
\newblock {\em {IEEE} Transactions on Pattern Analysis and Machine
  Intelligence}, 35(11):2765--2781, 2013.

\bibitem{Favaro:CVPR11}
P.~Favaro, R.~Vidal, and A.~Ravichandran.
\newblock A closed form solution to robust subspace estimation and clustering.
\newblock In {\em IEEE Conference on Computer Vision and Pattern Recognition},
  2011.

\bibitem{Huang:CVPR04-ED}
K.~Huang, Y.~Ma, and R.~Vidal.
\newblock Minimum effective dimension for mixtures of subspaces: A robust
  {GPCA} algorithm and its applications.
\newblock In {\em {IEEE} Conference on Computer Vision and Pattern
  Recognition}, volume~II, pages 631--638, 2004.

\bibitem{Kossovsky14}
A.~Kossovsky.
\newblock {\em Benford's Law : Theory, the General Law of Relative Quantities,
  and Forensic Fraud Detection Applications}.
\newblock World Scientific Publishing Company, 2014.

\bibitem{LeCun:1998}
Y.~LeCun, L.~Bottou, Y.~Bengio, and P.~Haffner.
\newblock Gradient-based learning applied to document recognition.
\newblock In {\em Proceedings of the IEEE}, pages 2278 -- 2324, 1998.

\bibitem{Liu:TPAMI13}
G.~Liu, Z.~Lin, S.~Yan, J.~Sun, and Y.~Ma.
\newblock Robust recovery of subspace structures by low-rank representation.
\newblock {\em IEEE Trans. Pattern Analysis and Machine Intelligence},
  35(1):171--184, Jan 2013.

\bibitem{Liu:ICML10}
G.~Liu, Z.~Lin, and Y.~Yu.
\newblock Robust subspace segmentation by low-rank representation.
\newblock In {\em International Conference on Machine Learning}, 2010.

\bibitem{Lu:ECCV12}
C.-Y. Lu, H.~Min, Z.-Q. Zhao, L.~Zhu, D.-S. Huang, and S.~Yan.
\newblock Robust and efficient subspace segmentation via least squares
  regression.
\newblock In {\em European Conference on Computer Vision}, 2012.

\bibitem{Ma:SIAM08}
Y.~Ma, A.~Yang, H.~Derksen, and R.~Fossum.
\newblock Estimation of subspace arrangements with applications in modeling and
  segmenting mixed data.
\newblock {\em SIAM Review}, 50(3):413--458, 2008.

\bibitem{Mcwilliams14subspace}
B.~McWilliams and G.~Montana.
\newblock Subspace clustering of high-dimensional data: a predictive approach.
\newblock {\em Data Mining and Knowledge Discovery}, 28(3):736--772, 2014.

\bibitem{Soltanolkotabi:AS12}
M.~Soltanolkotabi and E.~J. Cand\`es.
\newblock A geometric analysis of subspace clustering with outliers.
\newblock {\em Annals of Statistics}, 40(4):2195--2238, 2012.

\bibitem{Tipping-mixtures:99}
M.~Tipping and C.~Bishop.
\newblock Mixtures of probabilistic principal component analyzers.
\newblock {\em Neural Computation}, 11(2):443--482, 1999.

\bibitem{Tron:CVPR07}
R.~Tron and R.~Vidal.
\newblock A benchmark for the comparison of 3-{D} motion segmentation
  algorithms.
\newblock In {\em {IEEE} Conference on Computer Vision and Pattern
  Recognition}, 2007.

\bibitem{Tsakiris:Asilomar14}
M.~C. Tsakiris and R.~Vidal.
\newblock Abstract algebraic-geometric subspace clustering.
\newblock In {\em Proceedings of Asilomar Conference on Signals, Systems and
  Computers}, 2014.

\bibitem{Tsakiris:AASC15}
M.~C. Tsakiris and R.~Vidal.
\newblock Abstract algebraic subspace clustering.
\newblock {\em CoRR}, abs/1506.06289, 2015.

\bibitem{Tseng:JOTA00}
P.~Tseng.
\newblock Nearest $q$-flat to $m$ points.
\newblock {\em Journal of Optimization Theory and Applications},
  105(1):249--252, 2000.

\bibitem{Vidal:SPM11-SC}
R.~Vidal.
\newblock Subspace clustering.
\newblock {\em {IEEE} Signal Processing Magazine}, 28(3):52--68, March 2011.

\bibitem{Vidal:PRL14}
R.~Vidal and P.~Favaro.
\newblock Low rank subspace clustering {(LRSC)}.
\newblock {\em Pattern Recognition Letters}, 43:47--61, 2014.

\bibitem{Vidal:CVPR04-gpca}
R.~Vidal, Y.~Ma, and J.~Piazzi.
\newblock A new {GPCA} algorithm for clustering subspaces by fitting,
  differentiating and dividing polynomials.
\newblock In {\em {IEEE} Conference on Computer Vision and Pattern
  Recognition}, volume~I, pages 510--517, 2004.

\bibitem{Vidal:CVPR03-gpca}
R.~Vidal, Y.~Ma, and S.~Sastry.
\newblock {Generalized Principal Component Analysis (GPCA}).
\newblock In {\em {IEEE} Conference on Computer Vision and Pattern
  Recognition}, volume~I, pages 621--628, 2003.

\bibitem{Vidal:PAMI05}
R.~Vidal, Y.~Ma, and S.~Sastry.
\newblock {Generalized Principal Component Analysis (GPCA)}.
\newblock {\em {IEEE} Transactions on Pattern Analysis and Machine
  Intelligence}, 27(12):1--15, 2005.

\end{thebibliography}

\end{document}